\documentclass[letterpaper,10pt,conference]{ieeeconf}

\IEEEoverridecommandlockouts
\usepackage{amsmath,amssymb}
\usepackage{graphicx}
\usepackage{booktabs}
\usepackage[nocompress]{cite}
\usepackage{url}
\usepackage{xspace}

\newcommand{\relu}[1]{\left[#1\right]_{+}}

\title{Distributed Swarm Morphogenesis with Neural Cellular Automata}

\author{Ray Verma, Anthony Tzes, and Eliseo Ferrante 
\thanks{R. Verma and E. Ferrante are with the Department of Computer Science, New York University Abu Dhabi United Arab Emirates}
\thanks{A. Tzes is with the Department of Electrical Engineering, New York University Abu Dhabi, United Arab Emirates }%
}

\begin{document}
\maketitle

\begin{abstract}
Swarm morphogenesis requires robots to assemble desired shapes through local interactions. We present a modular framework integrating neural cellular automata (NCA) with existing site assignment and collision-aware control. Within this framework, we introduce a decentralized attachment-site discovery mechanism. Robots already in the formation update and share local NCA states to identify new attachment sites, enabling progressive growth without a predefined list of sites during deployment. To bridge the mismatch between conventional NCA growth and gradual physical assembly, we additionally propose offline sparse-retention fine-tuning that simulates partial attachment, together with a false-positive loss that penalizes invalid site activations. Simulations demonstrate formation of diverse lattice morphologies, scaling to several hundred robots, and substantial improvements in morphology accuracy from the proposed fine-tuning. Experiments with Crazyflie quadrotors validate the integrated execution pipeline using robot-relative observations reconstructed from external tracking.

\end{abstract}

\section{Introduction}
\label{sec:introduction}

Swarm morphogenesis aims to generate large-scale physical structures through local interactions among robots. Applications include coordinated aerial displays, distributed inspection, search, surveillance, and defence, where formations may need to scale or adapt across environments. Reliance on GPS, motion capture, or centralized estimation can limit applicability where signals or infrastructure are unavailable, including GNSS-denied environments; relative-sensing approaches instead use onboard vision~\cite{montijano2016vision}, ranging and odometry~\cite{guo2020uwb}, or distance-only measurements~\cite{suttner2018distance,lu2025concurrent}. In this paper, we consider local-relative sensing without globally referenced robot poses.

Distributed task-allocation methods assign tasks to robots~\cite{choi2009consensus}, while assignment-based shape-formation methods allocate or dynamically select robot goals~\cite{wang2020taskswapping,li2025distributed}. Assignment-free methods guide robots to fill specified shapes without explicit robot-to-site assignments~\cite{sun2023meanshift,zhang2024meanshift}. These approaches differ in how robots occupy the shape, but both assume its geometry is supplied beforehand. We instead consider \emph{distributed attachment-site discovery}. Essentially, we determine which physical attachment sites should exist as the morphology develops, so that a global structure can emerge from a local developmental rule trained offline, rather than an explicitly stored goal list. Existing assignment and control methods then coordinate robots to occupy these sites.

NCAs are suitable for distributed attachment-site discovery because local computations generate global structure without a central coordinator. Each cell updates its state from neighbors, enabling pattern growth and regeneration~\cite{mordvintsev2020growing}, with extensions to graphs and higher-resolution generation~\cite{grattarola2021learning,gala2024equivariant,pajouheshgar2026cells}. We use these states to discover physical attachment sites, establishing a first step toward neural swarm morphogenesis. 

Our framework separates candidate-site discovery from assignment and control, allowing the latter two components to be replaced as independent modules. We use the artificial-light-field method for candidate-site assignment and the constrained controller for robot motion, both from~\cite{li2025distributed}.

We distinguish offline NCA training on virtual grids from deployment, where robots assemble the desired morphology incrementally and asynchronously through robot-local NCA updates. Standard Growing NCA ~\cite{mordvintsev2020growing} training omits physical assembly constraints, potentially impairing candidate-site predictions. We introduce offline sparse-retention fine-tuning to address this mismatch.

Our three main contributions are:
\begin{itemize}
    \item Modular lattice-morphogenesis framework. We introduce a new framework for progressive assembly of a desired morphology through local robot interactions.
    \item Distributed attachment-site discovery method. Robots in the formation exchange and update local NCA states to identify candidate attachment sites for unattached robots. This allows the formation to develop progressively without providing robots with a complete list of attachment sites.
    \item Offline fine-tuning adapts a pretrained NCA to physical assembly using simulated partial attachment. An additional false-positive loss penalizes activation outside the desired morphology.
\end{itemize}

\section{Background and Related Work}
\label{sec:background}

We review prior work through four methodological building blocks: morphology representation, development, assembly, and localization and control.

Morphology representation concerns how the desired shape is encoded. Prior approaches use continuous regions~\cite{hsieh2008decentralized} or density fields~\cite{xiang2026density}, as well as discrete lattices~\cite{nie2023triangular} or graphs~\cite{li2019progressive}. These representations specify spatial relationships. We adopt square lattices for their well-defined attachment sites during progressive assembly.

Morphology development concerns the rules by which a desired structure takes shape. Progressive self-assembly methods prescribe local rules for adding robots~\cite{rubenstein2014programmable,li2019progressive,pan2024crystallization}; their growth is decentralized, but the rules are rigid. Gene-regulatory models~\cite{meng2013morphogenetic} and reaction--diffusion dynamics~\cite{slavkov2018morphogenesis} have also been used for morphogenesis in robot swarms. 
In a separate line of research, learned neural cellular automata (NCAs) have generated patterns in virtual environments~\cite{mordvintsev2020growing}, including graph-based extensions~\cite{grattarola2021learning,gala2024equivariant}. Their applications to robotics include regeneration of simulated soft robots~\cite{horibe2021regenerating} and shape classification in physical modular robots~\cite{walker2022physical}. We investigate the use of learned NCAs for discovering attachment sites during shape formation by independently mobile robots. To the best of our knowledge, previous NCA-based systems have not combined learned candidate-site discovery with progressive physical assembly by independently mobile robots.

Graph Neural Networks (GNNs) enable learned decentralized interactions through neighborhood message passing~\cite{jiang2023gnn}. Our NCA follows a similar principle but operates on a regular lattice rather than a general graph. Its shared symbolic frame preserves neighbor directions, allowing spatially ordered updates without requiring permutation-invariant aggregation or explicit edge features. We therefore formulate progressive attachment-site generation using the simpler NCA setting, without requiring the general GNN framework. 

Morphology assembly concerns how robots occupy a specified morphology. Distributed task allocation provides coordination mechanisms~\cite{choi2009consensus}, with methods including local task swapping~\cite{wang2020taskswapping}, dynamic robot-to-site assignment~\cite{li2025distributed}, and assignment-free movement~\cite{sun2023meanshift,zhang2024meanshift}. 

Localization and control concern how robots perceive and reach positions. Existing approaches use GPS or indoor positioning~\cite{yang2019indoor}, relative observations based on vision, ranging, odometry, or distance measurements~\cite{montijano2016vision,guo2020uwb,suttner2018distance,lu2025concurrent}, and learned decentralized motion control~\cite{jiang2023gnn}. Our implementation uses local relative observations and reuses the constrained controller from~\cite{li2025distributed}.

\begin{figure*}[t]
    \centering
    \includegraphics[width=\textwidth]{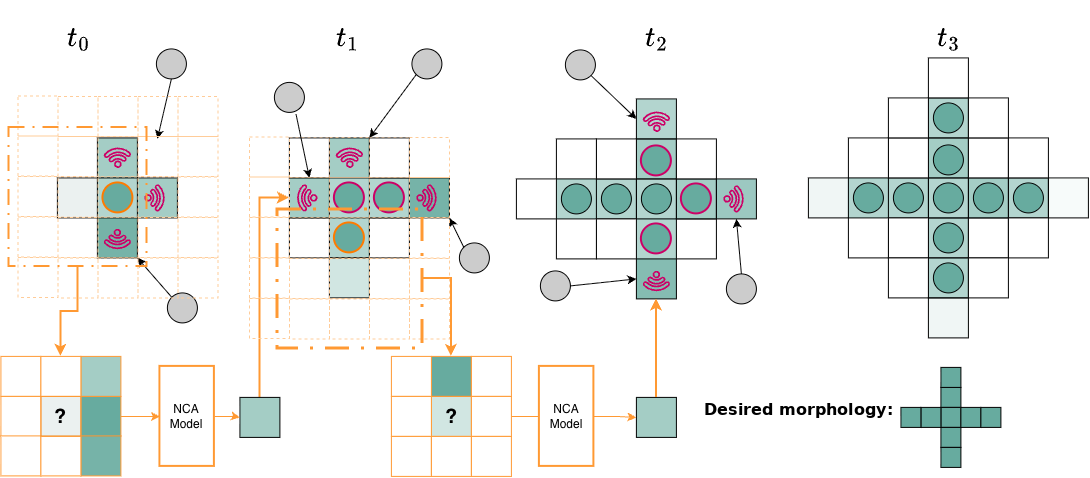}
    \caption{Swarm Morphogenesis over time. At each step $t_k$, Advertisers maintain local NCA state around their occupied symbolic cells, refresh a $5 \times 5$ neighborhood patch from observed attached neighbors, apply an NCA update, and advertise neighbouring unoccupied foreground cells as candidate attachment sites. Joiners can then occupy these advertised cells and become Advertisers, so the set of attached cells grows progressively toward the desired morphology.}
    \label{fig:nca_target_discovery}
\end{figure*}

\section{Methodology}
\label{sec:methodology}
Our method has two phases: offline training and deployment. During offline training, an NCA learns to reconstruct each desired morphology and is then fine-tuned to account for gradual robot assembly. During deployment, its weights remain fixed while robots exchange and update local states to discover candidate attachment sites. We first define the physical and symbolic setting, then explain local frame estimation, attachment-site discovery, assignment and control, and finally the offline training procedure.

\subsection{Problem Setting}
\label{subsec:problem}

We consider $N$ homogeneous holonomic robots progressively assembling a desired formation using only local sensing and communication. One robot is designated the initial \emph{seed} and remains stationary throughout assembly. An \emph{Advertiser} is a robot already attached to the formation that updates local NCA states and advertises candidate attachment sites; a \emph{Joiner} is an unattached robot that discovers and approaches these sites. After attaching, a Joiner becomes an Advertiser.

Possible attachment positions lie on a symbolic square lattice, with the seed at coordinate $(0,0)$ and adjacent sites separated physically by $\ell_{\mathrm{space}}$. The desired morphology is a finite, cardinally connected set of lattice cells $\mathcal{T}\subseteq\mathbb{Z}^2$ containing the seed, with at most one robot per cell and $|\mathcal{T}|\leq N$. In the reference morphology, cells belonging to $\mathcal{T}$ are \emph{foreground} and all other cells are \emph{background}. During assembly, an unoccupied cell may be predicted as foreground and be advertised for attachment, with Joiners moving towards these cells. The lattice specifies relative geometry; the complete desired morphology is available during offline training but is not provided to the robots during deployment.

Each robot measures its speed and the relative positions of sensed neighbors. Robots broadcast information such as their occupied lattice coordinates, local NCA states, and advertised candidate sites. Each Advertiser's NCA update depends on its local states and received neighbor information, not a shared global formation state. An Advertiser proposes only unoccupied sites adjacent to its own cell; Joiners discover advertised sites locally and can attach where the motion controller can reach the cell.

\subsection{Local Reference Frame Estimation}
\label{subsec:morphology_estimation}

Advertisers specify candidate sites in symbolic lattice coordinates. Joiners locate these sites by estimating a rigid transform into their own body frames using nearby Advertisers. For robot $i$, let $\mathcal{N}_i(t)$ be its sensed neighbors at time $t$, and let $\mathcal{A}_i(t)\subseteq\mathcal{N}_i(t)$ contain those already attached as Advertisers. For each such neighbor $k$, the Joiner observes body-frame position $p_k\in\mathbb{R}^2$ and receives its symbolic cell $c_k\in\mathbb{Z}^2$. Multiplying that cell by the prescribed spacing gives the corresponding lattice position $q_k=\ell_{\mathrm{space}}c_k$.

Robot $i$ estimates a rotation $R_i \in SO(2)$ and translation $t_i \in \mathbb{R}^2$ aligning scaled lattice coordinates with observed neighbour positions. This yields the rigid least-squares problem
\begin{equation}
    (R_i,t_i)
    =
    \underset{{R \in SO(2),\, t \in \mathbb{R}^2}}{\operatorname{argmin}}
    \sum_{k \in \mathcal{A}_i(t)}
    \left\|p_k-(Rq_k+t)\right\|_2^2 .
    \label{eq:morphology_fit}
\end{equation}
For any symbolic cell $c$, its estimated position in the body frame of robot $i$ is the vector $g_i(c)\in\mathbb{R}^2$, defined by
\begin{equation}
    g_i(c)=R_i\!\left(\ell_{\mathrm{space}}c\right)+t_i .
    \label{eq:morphology_projection}
\end{equation}

With only the seed as a reference, Eq.~\eqref{eq:morphology_fit} cannot determine the lattice orientation. The first Joiner attaches at a seed-advertised neighboring cell using its local frame; the physical direction from the seed to this second robot establishes the corresponding lattice axis and thereby fixes the orientation of the desired morphology. No global orientation is prescribed for the desired morphology. Once two attached robots with distinct known lattice coordinates are observable, subsequent Joiners can estimate the planar rigid transform; additional observations provide a least-squares fit, assuming that the observing robot can sense sufficient attached neighbors.

\subsection{Candidate-Site Discovery}
\label{subsec:attached_broadcast}

We use a Growing Neural Cellular Automaton (NCA) as the learned developmental model~\cite{mordvintsev2020growing}. A \emph{cell state} is a vector $x(c)\in\mathbb{R}^{C}$ associated with one symbolic cell $c$, comprising red, green, blue, and alpha (RGBA) values plus internal channels. A \emph{state patch} is a collection of such vectors. RGB describes appearance, and alpha indicates the NCA's predicted foreground membership. All Advertisers apply the same learned update rule $\mathcal{F}_{\theta}$, but each maintains its own local states. The rule computes each cell's update from its $3\times3$ neighborhood; its training is described in Section~\ref{sec:nca_training}.

Each Advertiser $i$ keeps a persistent $5\times5$ patch $X_i^t$ centered on its occupied cell $c_i$ at control step $t$. This patch provides local context needed for updating the the occupied center and its adjacent candidate cells. Figure~\ref{fig:nca_target_discovery} illustrates the process.

At each control step, the Advertiser refreshes $X_i^t$ with nearby occupied-cell states received from other Advertisers broadcasting their symbolic coordinates and NCA states. It then performs an NCA update on the patch
\begin{equation}
    {X}_i^{t+1}=\mathcal{F}_{\theta}({X}_i^t),
    \label{eq:distributed_nca_update}
\end{equation}
It examines the four cardinally adjacent cells of $c_i$. It advertises a cell $c$ only if the cell is locally known to be unoccupied and its updated alpha value $x^{alpha}(c) > \theta_{\alpha}$, where $\theta_{\alpha}$ is the foreground threshold. Each advertisement contains the candidate's symbolic coordinate and updated NCA state. 

A Joiner $j$ selects an advertised unoccupied cell $c_j$ and approaches it. When multiple Advertisers announce the same site, the Joiner averages their alpha values to evaluate the candidate. Upon attachment, it initializes the NCA state of its newly occupied cell from one of the advertised complete cell-state vectors. It then maintains its own local patch as an Advertiser and can advertise further candidate sites.

\subsection{Site Selection and Attachment}
\label{subsec:assignment_control}

Once candidate attachment sites are advertised, Joiners use the distributed artificial-light-field (ALF) assignment of Li \emph{et al.}~\cite{li2025distributed} to select candidates. The ALF computes local artificial-light intensities from observed robot positions, allowing Joiners to favor advertised sites with lower intensity in less crowded regions. 

The constrained controller of~\cite{li2025distributed} computes Joiner velocity commands toward the selected candidate site, balancing site tracking, collision avoidance, and velocity smoothing under speed and safety constraints. The ALF assignment method and constrained controller are replaceable components of our framework.

\subsection{NCA Training and Fine-Tuning}
\label{sec:nca_training}

We train the local operator $\mathcal{F}_{\theta}$ offline on the complete desired morphology. A standard Growing NCA first learns to reconstruct the reference image; sparse-retention fine-tuning then adapts it for gradual assembly and simulated attachment. 

\subsubsection{Growing NCA Training}
\label{subsec:nca_baseline}

The operator $\mathcal{F}_{\theta}$ follows Growing NCA~\cite{mordvintsev2020growing}: at each step, a convolution extracts features from each cell's $3\times3$ neighborhood, a shared network predicts a state change, and that change is added to the current state. A stochastic per-cell update mask emulates asynchronous updates during training. The same operator is used during deployment in Eq.~\eqref{eq:distributed_nca_update}. Starting from one living seed state, training unrolls $H$ update steps and optimizes the weights by backpropagation through time. Let $y$ be the reference RGBA image of the desired morphology and $x_H^{\mathrm{RGBA}}$ the NCA's predicted RGBA channels after the rollout. Their final-state reconstruction loss is
\begin{equation}
    \mathcal{L}_{\mathrm{RGBA}}
    =
    \operatorname{MSE}
    \left(
        x_H^{\mathrm{RGBA}},y
    \right).
    \label{eq:rgba_loss}
\end{equation}

\begin{figure*}[tb]
    \centering
    \includegraphics[width=1.0\textwidth]{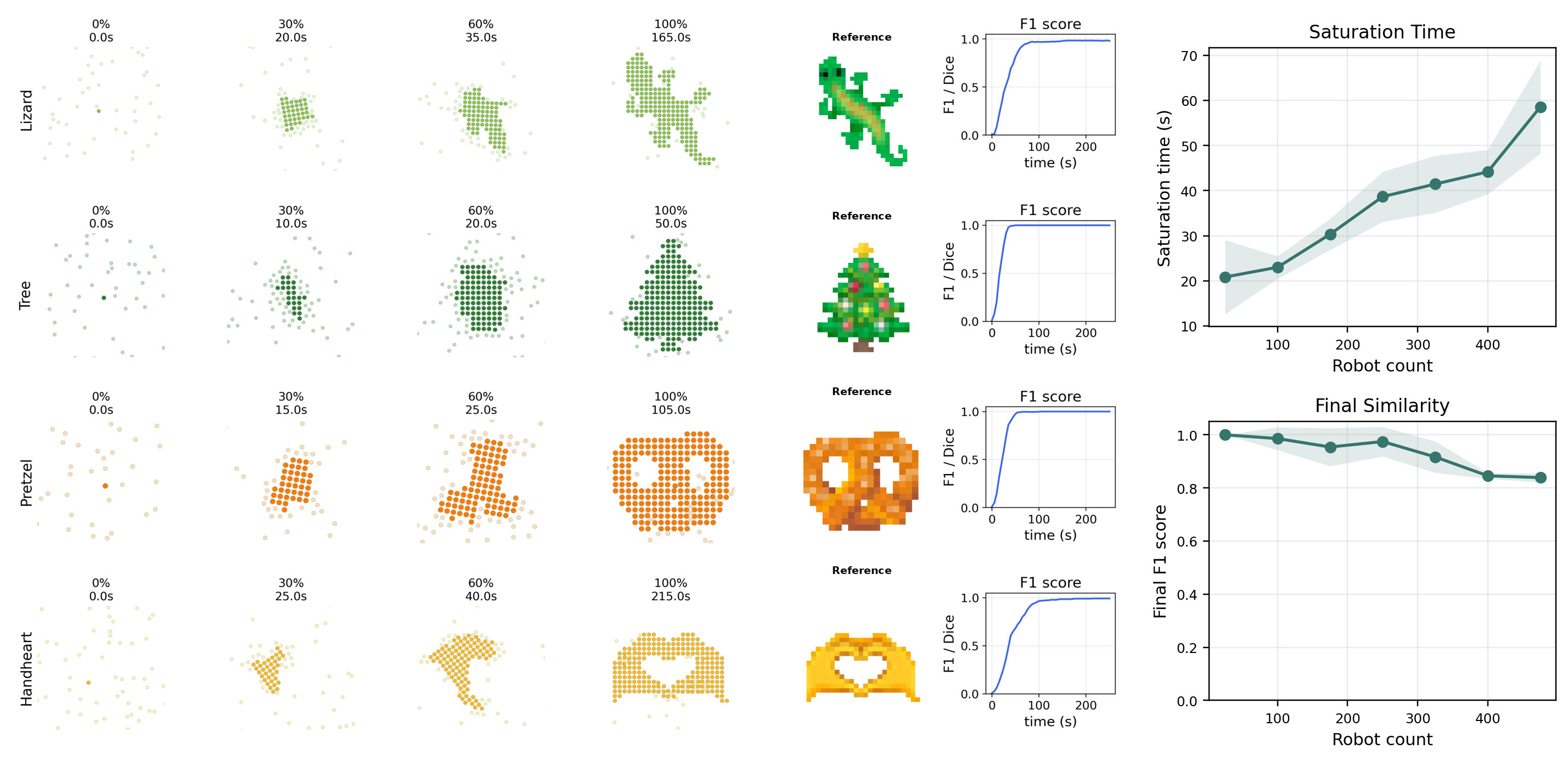}
    \caption{Formation progression and scalability. Left: representative assembly sequences for four desired morphologies using $250$ robots at $0\%$, $30\%$, $60\%$, and $100\%$ progression, with reference images and corresponding F1 accuracy trajectories over time. Right: saturation time and final F1 versus robot count for \texttt{Tree} morphology; shaded regions denote one standard deviation.}
    \label{fig:formation_scaling}
\end{figure*}

\subsubsection{Sparse-Retention Fine-Tuning}
\label{subsec:sparse_retention}

Ordinary NCA growth permits cells to activate temporarily or grow ahead of the structure. 
During physical assembly, however, only NCA states at \textit{occupied} robot cells are used to predict adjacent attachment sites, where each Advertiser updates only a local patch around its occupied cell as robots gradually join. 
Applying this restriction to a standard NCA may therefore expose it to unfamiliar developmental dynamics, and may produce unsuitable predictions from such partial, irregular states.
Sparse-retention fine-tuning exposes the model to this restricted regime so it can form the target progressively from the occupied frontier, improving deployment-time predictions.

Sparse-retention rollouts emulate gradual assembly: update virtual state, identify foreground cells adjacent to the occupied morphology, randomly retain some as newly attached, and prune unsupported state before continuing. Let $x_n$ be the state at step $n$, and $A_n\subseteq\mathbb{Z}^2$ the occupied lattice sites representing Advertisers. With $\operatorname{Adj}(A_n)$ denoting cardinally adjacent cells, the unoccupied frontier and eligible candidates are
\begin{align}
\mathcal{F}_n &= \operatorname{Adj}(A_n)\setminus A_n,\\
\mathcal{C}_n &= \{c\in\mathcal{F}_n: x_n^{\mathrm{alpha}}(c)\geq\theta_{\alpha}\},
\label{eq:sparse_candidates}
\end{align}
where $\theta_{\alpha}$ is the foreground retention threshold. To model slow and intermittent arrivals, a subset of candidates is sampled from the retention distribution $\Pi_{\mathrm{ret}}$:
\begin{equation}
K_n\sim\Pi_{\mathrm{ret}}(\cdot\mid\mathcal{C}_n),
\qquad
A_{n+1}=A_n\cup K_n,
\label{eq:retention_sampler}
\end{equation}
where $K_n\subseteq\mathcal{C}_n$ contains the newly retained candidates. 

To enforce the limited spatial extent of NCA state during physical assembly, virtual NCA states are kept only on occupied cells and cells immediately adjacent to them; all other states are set to zero:

\begin{equation}
x_n(c)\leftarrow 0
\quad\forall c\notin A_{n+1}\cup\operatorname{Adj}(A_{n+1}).
\label{eq:halo_prune}
\end{equation}
This permits nearby predictions while pruning distant activity, exposing the NCA to partial, irregular assemblies. 
Sparse-retention rollouts are also mixed with ordinary NCA rollouts to preserve original developmental dynamics.

\paragraph{False positive}
Because adjacency alone cannot exclude invalid sites, we additionally penalize foreground activation outside the desired morphology. Let $T(c)\in\{0,1\}$ indicate membership in the desired morphology and $\mathcal{M}=\{c:T(c)=0\}$ be background cells. Alpha above threshold $\tau_{\mathrm{fp}}$ incurs
\begin{equation}
\mathcal{L}_{\mathrm{FP}, n}
=\left\langle\relu{ x^{\mathrm{alpha}}_n (c)-\tau_{\mathrm{fp}}}^{2} \right\rangle_{c\in\mathcal{M}},
\label{eq:fp_loss}
\end{equation}
where $\left\langle\cdot\right\rangle_{\mathcal{M}}$ is the average over $\mathcal{M}$. Combining this penalty across the rollout with final RGBA reconstruction gives
\begin{equation}
\mathcal{L}_{\mathrm{SR}}
=\frac{\lambda_{\mathrm{FP}}}{H}\sum_{n=1}^{H}
\mathcal{L}_{\mathrm{FP},n}
+\lambda_{\mathrm{RGBA}}\mathcal{L}_{\mathrm{RGBA}}.
\label{eq:sparse_retention_loss}
\end{equation}
Thus fine-tuning encourages local physical growth while suppressing false-positive sites. The complete desired morphology and training losses are used only offline.

\section{Results}
\label{sec:results}


For each desired morphology, we train a dedicated NCA offline. We then conduct two simulation studies: (i) formation progression and scaling, to evaluate the framework's ability to assemble different morphologies and accommodate increasing swarm sizes; and (ii) a fine-tuning ablation, to assess the effects of sparse retention and false-positive loss on deployment accuracy. We also perform physical quadrotor experiments are presented in Section~\ref{subsec:hardware_validation}.

For offline training, a Growing NCA first learns to reconstruct a reference image and is subsequently fine-tuned using sparse retention. For an $N$ robot experiment, the reference image is scaled so that its foreground-cell count is closest to $N$. For fine-tuning, the retention distribution $\Pi_{ret}$ selects no candidate sites with probability $0.15$, all with probability $0.10$, and otherwise retains candidates independently with probability $\rho_n\sim\mathcal{U}(0.1,0.6)$. 

For all simulations, NCA weights remain fixed and robot-local updates are asynchronous. Initial robot positions are sampled uniformly in the arena. Results are computed over $30$ random seeds. Experimental parameters are reported in  Table~\ref{tab:main_parameters}.

\begin{table}[t]
\centering
\caption{Parameters for offline training and simulated deployment.}
\label{tab:main_parameters}
\scriptsize
\setlength{\tabcolsep}{5pt}
\begin{tabular}{ll}
\toprule
\textbf{Parameter} & \textbf{Value} \\
\midrule
State channels $C$ & $16$ \\
Foreground threshold $\theta_{\alpha}$ & $0.5$ \\
Training Rollout horizon $H$ & $[64,96)$ steps \\
False-activation threshold $\tau_{\mathrm{fp}}$ & $0.2$ \\
Loss weights $(\lambda_{\mathrm{FP}},\lambda_{\mathrm{RGBA}})$ & $(2.0,4.0)$ \\
Lattice spacing $\ell_{\mathrm{space}}$ & $0.5$ \\
\bottomrule
\end{tabular}
\end{table}

We measure morphology accuracy by the F1 score (Dice overlap) between the expected and occupied cells. Let $\mathcal{T}$ denote the foreground cells of the desired morphology and $\mathcal{A}(t)$ the cells occupied by robots at time $t$:
\begin{equation}
F_1(t)=\frac{2|\mathcal{A}(t)\cap\mathcal{T}|}{|\mathcal{A}(t)|+|\mathcal{T}|}.
\label{eq:morphology_f1}
\end{equation}
F1 equals $1$ when the occupied cells exactly match the desired morphology, and decreases when cells are missing or incorrectly occupied.

For an episode ending at time $T$, saturation time is the earliest time from which the $F_1$ score remains close to the final value:
\begin{equation*}
\begin{aligned}
t_{\mathrm{sat}}=\min_{t\in[0,T]:}\big\{
 |F_1(\tau)-F_1(T)|\leq\varepsilon_{\mathrm{sat}} ~;~ 
 \forall\tau\in[t,T]\big\}.
\end{aligned}
\label{eq:saturation_time}
\end{equation*}
with $\epsilon_{sat}$ such that the F1 score at $t_{\mathrm{sat}}$ is within $5\%$ of its terminal value $\epsilon_{sat} = 0.05 \times F_1(T)$.

\subsection{Simulation Results}
\label{subsec:simulation_results}

\paragraph{Formation and Scaling} 
We evaluate four morphologies (\texttt{Lizard}, \texttt{Tree}, \texttt{Pretzel}, and \texttt{Hand-heart}) using $250$ robots, and vary the robot count for \texttt{Tree}. 
Figure~\ref{fig:formation_scaling} contains representative formation rollouts and scaling results. Snapshot progression percentages are computed with respect to $F_1(t_{sat})$. 
Dark colored robots denote Advertisers, while light colored robots denote Joiners. Formations are rotated and scaled for visualization. 

The rollouts show that the local developmental mechanisms can form branched, compact, and holed lattice morphologies without a deployment-time goal list. Morphologies that expand broadly in multiple directions, such as Tree, form faster, whereas those with narrow paths or constrained branches, such as Lizard and Hand-heart, require longer formation times. Nonetheless, these runs converge towards F1 values close to $1$, confirming that the assembled formations increasingly match their desired morphologies. In the scaling study, $t_{\mathrm{sat}}$ grows approximately linearly with robot count, while the final F1 score only decreases slowly, indicating that morphologies remain stable across scale. The shaded regions show one standard deviation.

\begin{figure*}[tb]
    \centering
    \includegraphics[width=1.0\textwidth]{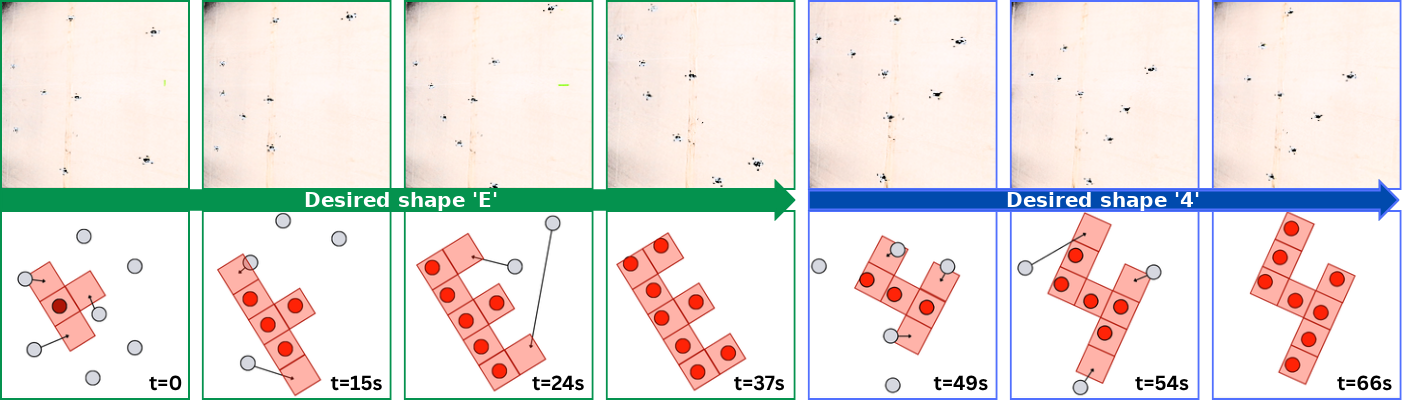}
    \caption{Real-robot experiments with eight Crazyflie quadrotors. Top: overhead snapshots during formation of \texttt{E} and \texttt{4}. Bottom: corresponding lattice-state visualizations, with gray and red circles denoting Joiners and Advertisers respectively, and colored cells indicating foreground morphology sites.}
    \label{fig:real_snapshots}
\end{figure*}
\paragraph{Ablation Study}
To evaluate the proposed developmental attachment-site generation mechanism independently of the assignment and control components, we use standard Growing NCA as the baseline and progressively introduce the proposed sparse-retention and false-positive suppression mechanisms. Table~\ref{tab:ablation} reports the deployment F1 score at $t_{sat}$ for these configurations on the \texttt{Tree} morphology with $250$ robots.

Sparse-retention increases the final F1 score relative to standard Growing NCA, demonstrating the benefit of training on partial assembly states. Adding the false-positive loss further improves mean F1 and significantly reduces variation across runs.
The ablation thereby provides direct empirical justification for the proposed fine-tuning procedure, that significantly  increases both performance and consistency.

\begin{table}[t]
\centering
\caption{Fine-tuning ablation.}
\label{tab:ablation}
\scriptsize
\setlength{\tabcolsep}{6pt}
\begin{tabular}{lcc}
\toprule
Training configuration & $F_1(t_{\mathrm{sat}})\uparrow$ & Std. \\
\midrule
Standard Growing NCA & $0.583$ & $0.107$ \\
\quad + Sparse-retention fine-tuning & $0.764$ & $0.141$ \\
\quad + \textbf{False-positive loss} & $\mathbf{0.988}$ & $\mathbf{0.004}$ \\
\bottomrule
\end{tabular}
\end{table}

\subsection{Hardware Validation}
\label{subsec:hardware_validation}

For hardware validation, we use eight Crazyflie quadrotors to assemble two desired lattice morphologies, \texttt{E} and \texttt{4}. Since the platform lacks the required relative-position sensing, Lighthouse tracking and centralized computation are used as an experimental interface to provide robot positions and reconstruct local-frame observations. These local observations are consumed by the proposed framework and return body-frame velocity commands. This setup preserves the algorithm's local information structure despite the platform's sensing limitations.

Figure~\ref{fig:real_snapshots} shows overhead snapshots and corresponding lattice-state visualizations during the formation of \emph{E} and \emph{4}. The sequences illustrate progressive attachment-site discovery and robot recruitment, resulting in the desired formations. These experiments demonstrate the execution of the proposed pipeline on physical robots despite localization noise, actuation error, and asynchronous motion


\section{Conclusion}
\label{sec:conclusion}

We presented a modular NCA-based framework for distributed swarm morphogenesis, integrating learned attachment-site discovery with existing assignment and collision-aware control. Robots progressively assemble desired formations without a predefined list of attachment sites during deployment. The ablation demonstrates that fine-tuning substantially improves morphology accuracy over standard Growing NCA, while the additional false-positive loss achieves high accuracy with consistent results. Simulations demonstrate diverse morphologies and scalability to large swarms, while Crazyflie experiments validate the integrated pipeline on physical robots.


\bibliographystyle{IEEEtran}
\bibliography{references}

\end{document}